\documentclass[a4paper]{article}
\usepackage{INTERSPEECH2022}
\usepackage{multirow}
\usepackage{caption}
\usepackage{subcaption}

\title{Depression detection in social media posts using affective and \\ social norm features}
\name{Ilias Triantafyllopoulos$^1$, Georgios Paraskevopoulos$^{1,2}$, Alexandros Potamianos$^1$}
\address{
  $^1$School of ECE, National Technical University of Athens, Greece\\
  $^2$Institute for Speech and Language Processing, Greece}
\email{hliastriant1@gmail.com, \{geopar,potam\}@central.ntua.gr}

\begin{document}

\maketitle
\begin{abstract}
We propose a deep architecture for depression detection from social media posts. 
The proposed architecture builds upon BERT to extract language representations from social media posts and combines these representations using an attentive bidirectional GRU network.
We incorporate affective information, by augmenting the text representations with features extracted from a pretrained emotion classifier.
Motivated by psychological literature we propose to incorporate profanity and morality features of posts and words in our architecture using a late fusion scheme.
Our analysis indicates that morality and profanity can be important features for depression detection.
We apply our model for depression detection on Reddit posts on the Pirina dataset, and further consider the setting of detecting depressed users, given multiple posts per user, proposed in the Reddit RSDD dataset.
The inclusion of the proposed features yields state-of-the-art results in both settings, namely $2.65\%$ and $6.73\%$ absolute improvement in F1 score respectively.

\end{abstract}
\noindent\textbf{Index Terms}: Depression detection, BERT, Feature fusion, Emotion recognition, profanity, morality 

\section{Introduction}

Depression is a mental health disorder which affects a large portion of society. More specifically, \cite{world2017depression} records 322 million people worldwide suffering from depression, which corresponds to 4.4\% of the world population. Interestingly, the number of depressed people increased by 18.4\% from 2005 to 2015 \cite{vos2016global}. 

In recent years, social media have become pervasive platforms, where people share information, opinions, thoughts and feelings.
For many people, social media platforms are places where they turn for support and self-disclose about their mental health. 
According to \cite{lin2016association} and \cite{keles2020systematic}, there is a close association between depressive young adults and the excessive use of social media. De Choudhury et al \cite{de2013predicting} find that depressed social media users have different online activity and behavior patterns than non-depressed users. These studies indicate the feasibility of depression detection from social media posts. Automatic depression detection could enable positive applications for the users, from suicide prevention to the development of effective, (semi-)automated mental health resources.

Psychological studies indicate that depressive disorders influence the affected people's language use.
In \cite{o2007empathy}, depression is considered as a disorder of ``concern for others'', as depressed individuals have often elevated levels of empathy. 
The authors bring evidence from neuroscience and psychology that demonstrate a connection between morality and depression. 
In \cite{bucur-etal-2021-exploratory}, the authors find a positive correlation between offensive language and depressed individuals in their social media posts. 
Vingerhoets et al \cite{vingerhoets2013swearing} argue that excessive swearing can lead to isolation, and therefore feelings of rejection and depressive disorder.
Emotive language is also correlated with depression, as mental health issues affect the emotional state of people. 
It is empirically established that depressed individuals express more negative thoughts, emotions and perspectives \cite{rude2004language,hamilton1983cognitive,xezonaki2020affective}.

Depression detection from social media can be performed either at the individual post level or at the user level, given a collection of posts by said user.
In \cite{nguyen2014affective}, authors classify depression-related LiveJournal posts, while in \cite{de2013predicting} authors focus on Twitter post classification.
In \cite{coppersmith2015clpsych}, a shared task for CLPsych 2015 is proposed for clinical diagnoses from Twitter posts.
In \cite{moreno2011feeling}, authors manually collect $200$ college students' posts from a one year timespan and perform user level depression classification. 
In \cite{li2015attitudes}, the authors analyze the social attitudes of users who indicate intention to commit suicide in public comments.


Linguistic features have been previously shown to improve depression detection models \cite{xezonaki2020affective,coppersmith2015adhd, lyons2018mental}.
One popular resource is the LIWC lexicon \cite{pennebaker2001linguistic}, which contains word scores across multiple linguistic (e.g. morality, functions etc) and affective dimensions. Specifically, \cite{coppersmith2015adhd} investigate the use of LIWC features for detecting a range of mental health conditions in Twitter posts.
\cite{lyons2018mental} perform a similar analysis in forum posts, both at the user and at the post level.
Xezonaki et al \cite{xezonaki2020affective} detect depression from transcribed clinical interviews, injecting affective features from lexica.
In \cite{inproceedings}, Yadav et al use emotion, sarcasm, personality and sentiment as features to identify medical conditions in online health fora, while in \cite{yadav-etal-2020-identifying} they explore the importance of figurative language for detecting depressive symptoms.
In \cite{amir2019mental}, the authors have utilized demographic features for depression and PTSD detection from Twitter data.



In this work, we consider depression detection in social media posts, both at the individual post level and at the user level.
We propose a hierarchical, two-level architecture based on BERT.
We combine BERT representations with features extracted from a pretrained emotion detector. 
Our key contribution is the inclusion of additional features related to the profanity and morality scores of individual posts and users, motivated by works in the psychology domain.
Our extensive data analysis and ablation studies indicate that emotion, profanity and morality can improve depression detection both at the post and at the user level.
The proposed architecture achieves state-of-the-art results for both tasks.
Our code is publicly available \footnote{https://github.com/IliasTriant/socialmediadepressiondetection}.


\section{Methodology}

\begin{figure}[t]
  \centering
  \includegraphics[width=\linewidth, height=8 cm]{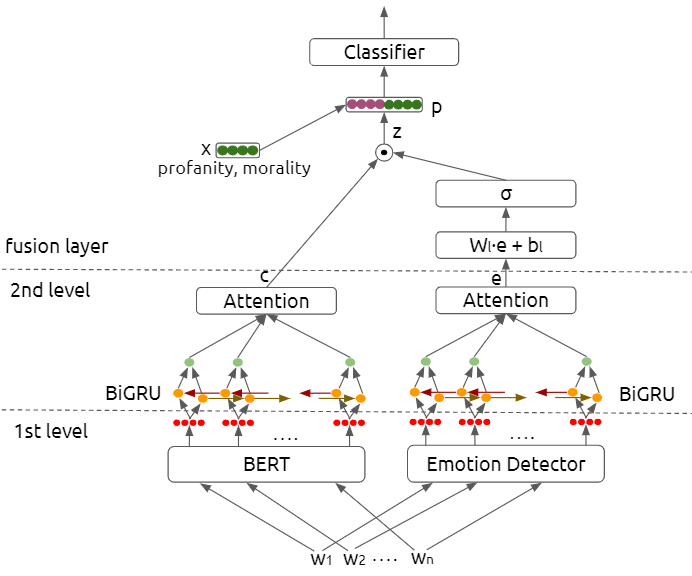}
  \caption{
  The proposed multi-level architecture.
  }
  \label{fig:architecture}
\end{figure}

Figure \ref{fig:architecture} illustrates an overview of the system architecture.
The architecture consists of three layers. At the first level we extract representations from BERT and the Emotion Detector.
At the second level we pass the BERT / Emotion Detector representations through an Attentive BiGRU network.
At the fusion layer, the BERT and Emotion representations are fused, along with the profanity and morality features.

\subsection{First Level}
 \noindent\textbf{Base Model:} We extract semantic representations from input text using a pretrained BERT model (\texttt{bert-base-uncased}) \cite{devlin-etal-2019-bert}.
 Assume a post $S^k_i = (w^k_{i,1}, w^k_{i,2}, ..., w^k_{i,n})$, where $w^k_{i,j}$ is the $j$-th word of the $i$-th post, $n$ is the maximum number of words and $k$ is an optional user annotation for the post (if we perform user level classification).
 We extract word representations $u^k_{i,j}$ from BERT ($f$) as in Eq.~\ref{eq1}.
\begin{equation}
  u^k_{i,j} = f(w^k_{i,1}, w^k_{i,2}, ..., w^k_{i,n})
  \label{eq1}
\end{equation}
In the case of post-level classification $u_{i,j}$ are passed to the BiGRU at the next architecture level.
In the case of user-level classification, given $m$ posts per user, we average $u^k_{i,j}$ to produce $u^k_i$, the $i$-th post representation for the $k$-th user. We pass then $u^k_i, i \in \{1, .., m \}$ posts, related to user $k$, to the BiGRU.



\noindent\textbf{Emotion Dectector:} We use a pretrained emotion detector, $g$, to extract affective information from the input text. In the post level classification, we extract the representation of the $j$-th word of the $i$-th post $v_{i,j}$ as shown in Eq.~\ref{eq2}.  
\begin{equation}
v_{i,j} = g(w_{i,j})
  \label{eq2}
\end{equation}
In the case of user level classification, the representation of the $i$-th post from the $k$-th user $v^k_i$ is produced from the emotion detector as in Eq.~\ref{eq3}.  
\begin{equation}
v^k_{i} = g(w^k_{i,1}, w^k_{i,2}, ..., w^k_{i,n})
  \label{eq3}
\end{equation}

\subsection{Second Level}

In the second level, we create the unified representation from the representations created in the first level.
As it is presented in Fig.~\ref{fig:architecture}, a Bi-directional Gated Recurrent Unit (Bi-GRU) network \cite{cho-etal-2014-learning} is used on top of BERT and the emotion detector. 
Let $\overrightarrow{h_i}$ be the $i$-th forward hidden state obtained by GRU associated with the BERT representations and $\overleftarrow{h_i}$ the corresponding backward hidden state. We concatenate $\overrightarrow{h_i}$ and $\overleftarrow{h_i}$ to obtain the hidden states $c_i$ for the BERT branch: 
\begin{equation}
  c_i = \overrightarrow{h_i} \mathbin\Vert \overleftarrow{h_i}
  \label{eq4}
\end{equation}

Similarly, let $\overrightarrow{k_i}$ and $\overleftarrow{k_i}$ be the $i$-th forward and backward hidden states of the BiGRU associated with the emotion detector. Then, we obtain $e_i$, the hidden states for the emotion branch, as follows: 
\begin{equation}
  e_i = \overrightarrow{k_i} \mathbin\Vert \overleftarrow{k_i}
  \label{eq5}
\end{equation}

We aggregate the hidden states $c_i$ and $e_i$ into the final representations $c$ and $e$ for the BERT and emotion branches respectively using an Attention Mechanism \cite{bahdanau2016neural}, on top of each of the respective GRUs. Given the hidden states $c_i$, we obtain the unified representation $c$ for the BERT branch  as the weighted sum of $c_i$ and the attention weights $a_i$:

\begin{equation}
  \begin{aligned}
t_i = q(c_i) \quad
\alpha_i = \frac{e^{t_i}}{\sum_i{e^{t_i}}} \quad
c = \sum_i {\alpha_i t_i}
\end{aligned}
  \label{eq6}
\end{equation}
where $q$ is a learnable mapping. We obtain the unified representation $e$ for the emotion branch in a similar manner.

\subsection{Fusion Layer}

In this architecture level, given the representations $c$ and $e$, we combine them, in order to extract the final representation. 
Inspired by \cite{margatina-etal-2019-attention}, we use a gating mechanism.
Specifically, a mask-vector is generated from $e$ (the representation associated with the emotion detector) with values between 0 and 1 and selects the salient dimensions of $c$ (the representation associated with BERT).
Concretely, 
\begin{equation}
z = c \odot \sigma(W_l \cdot e + b_l)
  \label{eq7}
\end{equation}
where $W_l, b_l$ are learnable parameters, $\sigma$ is the sigmoid function and $z$ is the final representation, which is fed into the classifier.

\subsection{Additional Features}

In the last level of our architecture, we add additional features, as shown in Fig.~\ref{fig:architecture}.
These features are chosen through our data analysis and they can be profanity, morality or both. They will be explained further in Section~\ref{analysis}.
We construct a vector $x$ that includes the features related to the second level entity. 
In the case of post classification, the entity is a post, whereas in the user classification it is a user. 

We integrate the $x$ vector to our architecture right before the classifier. 
Specifically, we concatenate, $\mathbin\Vert$, the $x$ vector with the representation extracted from the fusion layer, $z$ as in Eq.~\ref{eq8}. 

\begin{equation}
  p = x \mathbin\Vert z
  \label{eq8}
\end{equation}


\section{Datasets and Analysis} \label{analysis}

\begin{figure*}[t]
  \centering
  \includegraphics[width=.75\textwidth]{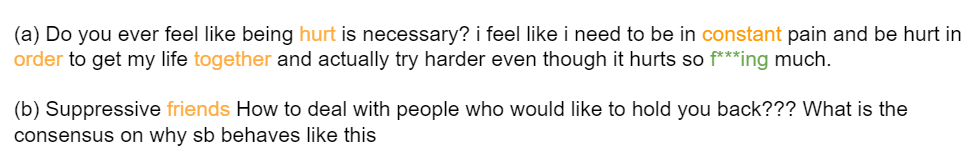}
  \caption{Examples of the Pirina dataset for (a) depression-indicative post, (b) standard post. Green: profane words, Orange: words related to social norms (morality) as found in the expanded LIWC.
  }
  \label{fig:samples}
\end{figure*}

\subsection{Datasets}

\textbf{Reddit Self-reported Depression Diagnosis (RSDD) dataset: } 
RSDD dataset \cite{yates-etal-2017-depression} consists of Reddit posts. Multiple posts are included per user, and users are labeled as depressed when the poster self-reports that they suffer from depression.
RSDD consists of training, validation and testing data and each of them contains approximately $3,000$ depressed users and $35,000$ matched control users. 
Posts that contain depression-related keywords or are published in mental health-related subreddits have been removed.
Every user in the dataset has approximately $900$ posts. 
In Table~\ref{tab:rsddstats}, we present the average number of posts per user, as well as the average number of words that are used in the posts for both classes, in the training set. 
We notice that depressed users post twice as much as the non-depressed and they use more words per post in a rate of $5:3$.\footnote{As per the dataset sharing agreement, we are not allowed to share samples from the RSDD dataset.}

\begin{table}[th]
  \caption{Datasets statistics for posts by depressed (D) and not depressed (ND) individuals.}
  \label{tab:rsddstats}
  \centering
  \begin{tabular}{l l l l  }
    \toprule
    \multicolumn{1}{c}{} &
    \multicolumn{1}{c}{Feature} & 
    \multicolumn{1}{c}{ND} & 
    \multicolumn{1}{c}{D}  \\
    \midrule
    \multirow{2}{*}{RSDD} & Avg number of posts/user  & 895.43 &  1739.54         \\
     & Avg number of words/post   &  24.38  & 40.80          \\
    Pirina & Avg number of words/post   &  297.08 & 182.42\\
    \bottomrule
  \end{tabular}
  
\end{table}

\smallskip

\noindent \textbf{Pirina: } Pirina dataset was built by Inna Pirina et al. \cite{pirina2018identifying} and contains $1,841$ posts extracted from users in Reddit. 
Among these posts, $1293$ are annotated as ``depression-indicative'' posts, whereas the remaining $548$ are annotated as ``standard'' posts.
In Fig.~\ref{fig:samples}, we present one sample of the depression-indicative posts and one standard post.
We see that the depression-indicative post is more emotional, as it contains more affective words, e.g. ``hurt'' and ``pain''.
Moreover, in Table \ref{tab:rsddstats}, we provide statistics about the average number of words per posts that are used in both classes. 
We observe that, in contrast with the RSDD Dataset, standard posts include more words than the depression-indicative posts.

  

\smallskip

\noindent \textbf{EmoBank-2017: } 
The emotion detector has been trained on the EmoBank-2017 dataset \cite{buechel2017emobank}, examining the posts on fine-grained emotions (``valence'', ``arousal'' and ``dominance''). 
EmoBank-2017 dataset consists of 10,062 posts, annotated in terms of valence, arousal and dominance in a 5-point scale. 
For the architecture of the emotion detector, we leverage a CNN network, similar to the one proposed in \cite{chen2015convolutional}.



\subsection{Profanity} 

There are signs that indicate higher usage of offensive language from depressed individuals.
We decide to measure the profanity score of a sentence, e.g. its profanity probability. 
To this purpose, we use the profanity-check tool, as proposed by \cite{profanity}. 
For the RSDD, we associate a profanity score with each user and use it as feature. This score is calculated as the average profanity score of user's posts. 
For the Pirina dataset, we use the post's profanity score as feature. 
The average profanity scores of depressed and non-depressed users/posts for the RSDD/Pirina dataset are presented in Table~\ref{tab:profanity}. 
For both datasets we observe higher profanity scores for the depressed category, and especially in the case of the Pirina. 
In the first sample of Fig.~\ref{fig:samples}, we see that the profane word ``f***ing'' determines the high profanity score of the sentence ($0.949$), while the second sentence does not contain a profane word and its score is low ($0.021$). 

\begin{table}[th]
  \caption{Average profanity scores per class.}
  \label{tab:profanity}
  \centering
  \begin{tabular}{ l l l  }
    \toprule
    \multicolumn{1}{c}{} & 
    \multicolumn{1}{c}{ND} & 
    \multicolumn{1}{c}{D}  \\
    \midrule
    RSDD  & 0.136 &  0.150          \\
    Pirina   &  0.166  & 0.308          \\
    \bottomrule
  \end{tabular}
  
\end{table}

\subsection{Social Norms} 
\begin{figure}
\centering
\begin{subfigure}{0.25\textwidth}
  \centering
  \includegraphics[width=\linewidth]{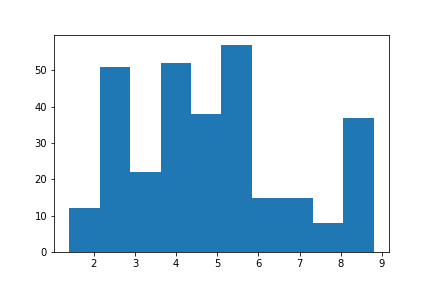}
 
  \end{subfigure}%
\begin{subfigure}{0.25\textwidth}
 \centering
  \includegraphics[width=\linewidth]{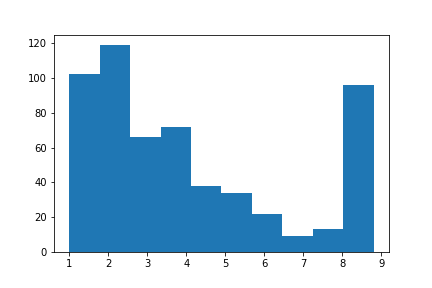}
\end{subfigure}
\caption{Histograms from the Pirina dataset, indicating the number of words as a function of the score in the Harm/Care (1-9) scale.ND: left  and D: right.}
 \label{fig:hist}
\end{figure}

To extract social norm features, we leverage the moral dimensions of Moral Foundations Dictionary (MFD) \cite{graham2009liberals} that was used together with LIWC \cite{pennebaker2001linguistic}, and its expansion \cite{araque2020moralstrength}. 
In particular, \cite{araque2020moralstrength} introduce the ``moral strength'' value, by providing a score in a $9$-point scale, which declares the moral valence of a word that belongs to a moral category. 
We focus on $5$ moral dimensions of the expanded MFD: Harm/Care, Cheating/Fairness, Betrayal/Loyalty, Subversion/Authority, Degradation/Purity. When a word is scored with 9, it belongs exclusively to the good option of the dimension, e.g. to ``Care'' if it is in Care/Harm, while when it is scored with 1, it belongs to the bad option, e.g. to ``Harm''. 
In the samples of Fig.~\ref{fig:samples}, we observe that the depression-indicative post has more words that belong to the moral vocabulary than the standard post.
In Table~\ref{tab:morality} we see the statistics of two features that are developed, in the RSDD dataset for both classes. These features are the average moral strength of the user and the percentage of user's posts that contain at least one word, which belongs to this moral dimension. We see that the main difference concerns the second feature, concluding that depressed users use, indeed, more moral vocabulary than the non-depressed. These are concatenated as global features. 
In Fig.~\ref{fig:hist} we see the overall histograms of the dimension Care/Harm for the Pirina dataset. 
We notice that depression-indicative posts contain more words that belong to the Care/Harm dimension, as also that the distribution on this dimension differs. 
More words tend to extreme values (near $1$ and $9$), while the standard posts are concentrated mainly on values between $4$ and $6$. 
Both differences are remarked on all dimensions.


\begin{table}[th]

  \caption{Average values of selected moral dimensions and the percentage of posts containing language related to these dimensions per class in the RSDD dataset.}
  \label{tab:morality}
  \centering
  \small
  \begin{tabular}{ l l l l }
    \toprule
    \multicolumn{1}{c}{Moral Dimension} & 
    \multicolumn{1}{c}{feature} & 
    \multicolumn{1}{c}{ND} & 
    \multicolumn{1}{c}{D}  \\
    \midrule
    \multirow{2}{*}{Harm/Care}& average & 4.07 & 4.26 \\
    & \% of posts & 6.42\% & 10.23\% \\
    \multirow{2}{*}{Cheating/Fairness}& average & 7.32 & 7.19 \\
    & \% of posts & 6.47\% & 10.45\% \\
    \multirow{2}{*}{Betrayal/Loyalty}& average & 6.32 & 6.47 \\
    & \% of posts & 4.72\% & 8.13\% \\
    \multirow{2}{*}{Subversion/Authority}& average & 6.36 & 6.36 \\
    & \% of posts & 6.50\% & 11.16\% \\
    \multirow{2}{*}{Degradation/Purity}& average & 6.73 & 6.64 \\
    & \% of posts & 8.20\% & 12.22\% \\
    \bottomrule
  \end{tabular}
  
\end{table}

\section{Experimental Setup} 

For user-level classification in the RSDD dataset, we extract post representations at the first level of the proposed architecture, and at the second level we combine post representations into user representations. 
For each user, we utilize $600$ posts. 
The models are trained for $10$ epochs using Adam with learning rate $10^{-3}$ and Cross Entropy Loss, with class weight ratio $1:7$. 

\smallskip

For post-level classification in the Pirina dataset, at the first level of our architecture we extract word representations, while at the second level we combine the word representations to obtain post representations.
We evaluate the model using $10$-fold cross-validation on $90/10$ train/test splits, following the state-of-the-art. 
For each fold we train a model for $5$ epochs using Cross entropy loss and Adam with learning rate  $5 \cdot 10^{-5}$. 

For all models we use Dropout $0.2$. All Bi-GRUs have $128$ hidden size.
We use PyTorch \cite{paszke2017automatic} in our implementation. 

For both datasets, we compare the performance of the BERT-only baseline (B) with the models that take into consideration the emotion detector (E), profanity (P) and morality (M). 
Thus, we experiment with (B+E) and then we add profanity (B+E+P), morality (B+E+M) and both (B+E+P+M). 
For both datasets, we report F1-score (F1), Recall (Re) and Precision (Pr). 
For the Pirina dataset, we also report Accuracy (Acc).

\section{Results and Discussion}

In Table~\ref{tab:RSDD}, we present the results of our models in the RSDD Dataset compared with the best models of \cite{rao2020mgl} and \cite{rao2020knowledge}. 
B+E model achieves a better F1-score and Recall than the previous best models, while adding profanity (B+E+P) we reach at the best overall improvement. 
B+E has approximately $24\%$ higher Recall than B, as more depressed users were classified as depressed.
On the contrary, the Precision decreases when we add the emotion detector. 
This can be explained by the fact that the model draws strong associations between the expression of negative emotions and depression.
By integrating profanity (B+E+P) and morality (B+E+M) we observe further improvements in performance. Furthermore, we observe more balanced precision and recall scores.
Surprisingly, we observe that adding both Profanity and Morality (B+E+P+M) does not improve the results further. 
This issue is attributed to difficulties in hyperparameter tuning and the simplicity of the fusion method. 

\begin{table}[th]
  \caption{Results of different architectures on the RSDD Dataset.}
  \label{tab:RSDD}
  \centering
  \begin{tabular}{ l l l l }
    \toprule
    \multicolumn{1}{c}{\textbf{Model}} & 
    \multicolumn{1}{c}{\textbf{Pr}} & 
    \multicolumn{1}{c}{\textbf{Re}} & 
    \multicolumn{1}{c}{\textbf{F1}} \\
    \midrule
    SGL-CNN \cite{rao2020mgl}&  51 & 56 & 53            \\
    MGL-CNN \cite{rao2020mgl}&  63 & 48 & 54            \\
    KFB-BiGRU-Att \cite{rao2020knowledge}&  57 & 51 & 54       \\
    KFB-BiGRU-Att-AdaBoost \cite{rao2020knowledge}&  58 & 54 & 56             \\
    B & 64.03 & 53.16 & 58.09 \\
    B + E  & 53.86 & \textbf{65.90} & 59.27           \\
    B + E + P & 64.41 & 61.14 & \textbf{62.73}             \\
    B + E + M & 63.50 & 60.07 & 61.74           \\
    B + E + P + M & \textbf{71.37} & 54.56 & 61.84 \\
    \bottomrule
  \end{tabular}
  
\end{table}

In Table~\ref{tab:Pirina} we present the results of our models in the Pirina Dataset compared with the best models of \cite{tadesse2019detection} and \cite{ren2021depression}. 
Here, also, emotion, profanity and morality improve the overall performance. 
The combination of BERT, emotion and profanity (B+E+P) achieves the best overall results, similar to the RSDD.

Overall, we note that affective and social norms features improve performance, both in the case of user-level classification in the RSDD dataset and the case of post-level classification in the Pirina dataset. 
We also note that this improvement is observed both in the case of a large dataset (RSDD), where approximately $28$ million posts are used for training, and a very small dataset (Pirina) with under $2,000$ posts.

\begin{table}[th]
  \caption{Results of different architectures on the Pirina Dataset.}
  \label{tab:Pirina}
  \centering
  \begin{tabular}{ l l l l l}
    \toprule
    \multicolumn{1}{c}{\textbf{Model}} & 
    \multicolumn{1}{c}{\textbf{Acc}} & 
    \multicolumn{1}{c}{\textbf{F1}} & 
    \multicolumn{1}{c}{\textbf{Pr}} & 
    \multicolumn{1}{c}{\textbf{Re}} \\
    \midrule
     LIWC+LDA+bg \cite{tadesse2019detection}& 91 & 93 & 90 & 92           \\
    EAN  \cite{ren2021depression}& 91.30& - & 91.91 & -       \\
    B & 91.89 & 93.93 & 94.38 & 93.93 \\
    B + E   & 92.07  & 94.42 & 93.95 & 94.90            \\
    B + E + P &  \textbf{93.87} & \textbf{95.65}  & \textbf{95.16} & \textbf{96.14}            \\
    B + E + M  &  92.79 & 94.97 & 94.03 &  95.94          \\
    B + E + P + M & 92.61 & 94.80 & 94.93 &  94.71        \\
    \bottomrule
  \end{tabular}
  
\end{table}

\section{Conclusions}

In this work we apply affective and social norm features for the task of depression detection from social media posts.
The affective features are extracted from a pretrained emotion detector and fused with semantic representations, extracted from BERT.
We fuse the combined BERT and affective representations with social norm features related to profanity and morality dimensions in a hierarchical architecture.
The inclusion of social norm features is supported by our data analysis, which indicates their importance for depression detection.
The profanity scores are found to be especially important.
Interestingly, social norm features improve results irrespective of dataset size.
The proposed feature combination and architecture is evaluated for post-level and user-level depression detection, yielding state-of-the-art results in the RSDD and Pirina datasets.
In the future we plan to explore more elaborate fusion techniques for inclusion of social norm features.
Furthermore, we plan to explore probing techniques and the interpretability of proposed architectures.



\bibliographystyle{IEEEtran}

\bibliography{mybib}


\end{document}